\documentclass[letterpaper,12pt]{article}

\usepackage{tabularx,ragged2e,booktabs,caption}
\usepackage{graphicx} 
\usepackage{subfigure} 
\usepackage[export]{adjustbox}
\graphicspath{ {C:/Users/Mark/Documents/Thesis/MarkHarmonPapers/Figures/} }
\usepackage[authoryear,comma,longnamesfirst,sectionbib]{natbib}
\usepackage{algorithm}
\usepackage{algorithmic}
\usepackage{enumerate}
\usepackage{hyperref,mathtools}
\usepackage{breakcites}

\usepackage{blindtext}
\usepackage{booktabs}
\usepackage{amssymb}
\usepackage{amsmath}
\DeclareMathOperator*{\argmax}{arg\,max}
\title{Dynamic Prediction Length for Time Series with Sequence to Sequence Networks }

\author
{Mark Harmon,$^{1}$ Diego Klabjan$^{2}$\\\\
\normalsize{$^{1}$ Department of Engineering Sciences and Applied Mathematics}\\
\normalsize{$^{2}$ Department of Industrial Engineering and Management Sciences}\\
\normalsize{Northwestern University}\\
}
\date{}
\begin{document}
\maketitle

\begin{abstract}

Recurrent neural networks and sequence to sequence models require a predetermined length for prediction output length.  Our model addresses this by allowing the network to predict a variable length output in inference.  A new loss function with a tailored gradient computation is developed that trades off prediction accuracy and output length.  The model utilizes a function to determine whether a particular output at a time should be evaluated or not given a predetermined threshold.  We evaluate the model on the problem of predicting the prices of securities.  We find that the model makes longer predictions for more stable securities and it naturally balances prediction accuracy and length.

\end{abstract}
\section{Introduction}

Recurrent neural networks are very popular and effective at solving difficult sequence problems such as language translation, creation of artificial music, and video prediction.  New architectures, such as Sequence to Sequences networks by \citet{sutskever2014sequence} and Memory Networks by \citet{sukhbaatar2015end} are used to solve problems in language translation and answer questions using a large memory bank.  However, these problems generally have training data with given sequence outputs (for example, a model translating a sentence from English to Spanish).  Because input and output sequences are known a priori for these problems, it is possible to solve them with a fixed model architecture.   

A fixed model architecture is effective for sequences, but there are a number of problems related to multiple time series datasets that do not have a natural sequence size.  For example, a company may wish to predict the number of products to be shipped out for sale based upon customer demand.  Each product has a different amount of demand volatility, which can make an enormous difference in how far in advance they are willing to predict the demand of a product.  In this case, it would be extremely useful to have a model that can balance F1 score and the number of future predictions based upon a product's base demand.  

Another example, which we explore in this work, is financial security price prediction.  Some securities are extremely volatile, which makes prediction for longer times highly inaccurate.  On the other hand, low volatility securities are easier to predict further into the future.  

The biggest problem in multiple time series predictions when it comes to dynamic prediction length is that the training data does not exhibit output sequences of various length.  For this reason, a different model is required. In multiple time series, input sequences can be naturally created for example by a fixed-size sliding window.  However, the length of the output sequences can be dynamic since typically there is flexibility in how far to predict in the future.  In inference, we allow our prediction model to generate a different number of predictions depending on the current input sequence as well as a different number of predictions per time series.  The number of predictions the model generates depends on a thresholding function that determines the model's confidence of that particular output.  If the confidence is too low, we no longer consider the predictions our model generates for that particular sample.  

The main objective of our study is to create a prediction model that balances F1 score and predicting into the future.  The main challenge is the fact that samples which are taken from infinite time series do not naturally contain dynamic length predictions.  This aspect requires a different loss function that includes the notion of confidence and tailored computation of gradients on different length output sequences.  We create a model architecture relying on a novel loss function that allows for a dynamic number of output predictions.  We explain the ideas and concepts by utilizing predictions of several correlated financial securities.  In this case, rather than having to adjust the predictive length manually depending on market volatility, the model learns how far in advance it can confidently predict during training.  For security prediction, a model that is not limited to a fixed number of output predictions can provide much more robust price predictions.  For example, we expect that some security $j$ with high volatility during the training phase should result in fewer predictions.  On the other hand, if the same security $j$ is trained during a low volatility period, we expect the model to generate more predictions.  Clearly, a dynamic model that can easily adjust to the current training environment of a security can provide huge benefits.  In inference the model provides a natural way to stop generating output predictions.

Our work contains two main contributions.  First, we create a way to measure the confidence of a model's prediction without having to rely on Bayesian statistics.  Second, our model is the first of its kind to allow for dynamic prediction length with sequence to sequence networks.  Along the way, we have to tailor gradient computation.  

In our study, we use two financial security datasets which consist of several years of tick prices.  One contains five distinct securities and the other contains twenty-two different securities.  We find that our new architecture successfully balances prediction F1 score and the number of future predictions.  In addition, our architecture uses different prediction lengths at different times for each security due to stochastic drift between training and test sets.  The best dynamic output prediction length model is a sequence to sequence network which earns an F1 score of 0.503 in contrast to a traditional LSTM structure that only gets an F1 score of 0.209 for a single prediction.
 
In Section 4.2, we review two main subjects related to our work.  First, we inspect studies within the realm of deep learning related to our new model architecture.  Second, we analyze other work on predicting financial securities with a focus on machine learning and deep learning methods.  In Section 4.3, we present the dynamic prediction length model while in Section 4.4 we present a computational study based on securities.

\section{Related Work}

Similar to our concept of dynamic output prediction, Pointer Networks by \citet{vinyals2015pointer} are used for problems such as sorting variable sized sequences.  They use an attention mechanism that points to a particular part of the input sequence that is used as the next output.  Although this architecture can allow for variable input sizes, the output size is constrained to be the same size as the input.  Our model allows for any size output (unrelated to the input size) up to some arbitrary maximum size.  Pointer networks are also not applicable to our case since output is not a specific part of input.

\citet{graves2016adaptive} introduces adaptive computation time (ACT) for recurrent networks.  The author creates an additional metric that allows the network to continue ``pondering" the input through additional computation.  We can think of ACT as a model that in each time has a dynamic number of stacked LSTM or GRU cells.  We considered using ACT along with our architecture, but it increases the computational complexity by recalculating the feed-forward step a number of times.  Basing the stopping decision with respect to the natural choice of the output time requiring substantial computational time, does not work when most times a single layer is needed.  In time series with a walk forward strategy, smaller less complex models are more effective.  This renders ACT not appropriate.  In addition, online algorithms need to be agile, and adding computational time would be a detriment to a model.  Therefore, we choose to use traditional LSTM architectures that train much faster.

To achieve better training and dynamic output sizes, we added an additional term to the loss function and modify the measure of predictions.  Although there have been many new architectures such as Residual Networks by \citet{he2016deep}, Memory Networks by \citet{sukhbaatar2015end}, and Neural Turing Machines by \citet{graves2014neural}, none of these incorporate a new loss function with their architecture.  This tendency of not creating new loss functions is noted by \citet{janocha2017loss}. The authors explain that although there is a lot of work in neural network activations, optimization, and architecture, the loss function used for nearly all neural networks is a combination of log loss and L1/L2 norms.  \citet{janocha2017loss} show that functions that were previously deemed to be inferior in deep learning can be more robust than log loss for classification problems.  Therefore, it is important for studies to continue exploring loss functions to increase the network performance and to create a variety of new models.

There are some studies that significantly change the loss function in deep learning.  For example, GAN's by \citet{goodfellow2014generative} are immensely popular and design a loss function for their specific problem and architecture to balance generation and classification.  ACT by \citet{graves2016adaptive} also uses a unique addition to the loss function so that their network does not ``ponder" on the input for too long.  There is growing momentum for using the Wasserstein metric as the loss function as seen in the work by \citet{frogner2015learning}.  The Wasserstein metric is even successfully being used in GAN's in the work of \citet{arjovsky2017wasserstein}.  We expand the volume of work in this area by developing a loss function that encourages a model to have a dynamic output length at prediction.  In contrast to loss functions specific to problem type, our architecture can be used with any recurrent neural network architecture.

Next, we focus on works on predicting financial securities with deep networks.  Most work utilizing deep neural networks focuses on applying other forms of data for prediction, such as news about financial markets or specific companies.  \citet{chong2017deep} review many of the prediction methods commonly used for security prediction and the predicted outcome.  \citet{niaki2013forecasting}, focuses on predicting upward and downward movement of S\&P 500 with a deep feed-forward network.  \citet{ding2015deep} use historical pricing data in combination with financial news data with a deep feed forward network.  \citet{krauss2017deep} utilizes both random forests and deep feed-forward networks to find statistical arbitrage on the S\&P 500.  In contrast to the aforementioned models, we predict significant movement in prices, utilize temporal models, and predict multiple securities with a single model.  

\citet{sirignano2016deep} uses deep learning directly on financial security price data as well.  He utilizes limit order book data with multiple ask/bid prices for each security to predict the change in the spread.  In addition, he uses a deep feed-forward network and a separate model for the prediction of each security.   In contrast, we predict all securities with one model and apply recurrent neural networks in addition to feed forward networks.  Another similar work, \citet{dixon2016classification}, uses deep feed-forward networks for prediction directly on security prices.  In contrast, our study applies recurrent and convolutional recurrent models in addition to the baseline feed forward network.  

There is also work on security prediction beyond the standard feed-forward network.

\noindent \citet{borovykh2017conditional} uses convolutional neural networks over the security time series to make predictions. Others, such as \citet{bao2017deep}, use stacked auto-encoders and wavelet transformations to form an embedding of financial data, and feed this into an RNN model for prediction. \citet{chen2015lstm} use an RNN model on opening and closing security prices on the Chinese market with seven classification categories.  \citet{akita2016deep} use both textual and price information to predict a security price.  In contrast to these works, our model consists only of security price data.  Also, we apply sequence to sequence and convolutional LSTM models in addition to having a dynamic output length for security predictions.

\section{Model}

In this section, we explain the architecture of our model that predicts a dynamic number of outputs.  The model uses a sequence to sequence (Seq2Seq) network in combination with our new proposed loss function.  For details on Seq2Seq networks, please see the original paper by \citet{sutskever2014sequence}.  We also use attention in our model which is explained in the paper by \citet{bahdanau2014neural}.
 
In addition, we use teacher forcing.  We found that the first input to the decoder has a great impact on accuracy.  By far the best performing first decoder input is the ground truth associated with the last encoder input.

For a general Seq2Seq network, let $\theta$ describe the trainable weights, $X$ be our input sample, and $f_{t}^{q} (X,\theta,f_{i<t})$ be the final output via a softmax function at time $t$ for time series $q$, $q=1...Q$.  In the presence of several time series, each time series requires a separate softmax.  The general formulation for a Seq2Seq network with Kullback-Leibler divergence is the following with true labels $Y=\{Y_{t}^{q}\}_t$: $$\min_\theta E_{(X,Y)} \sum_{q=1}^Q  \sum_{t=1}^T KL(Y_{t}^{q} || f_{t}^{q} (X;\theta; f_{i<t} )).$$  $$\text{Here} \quad f_{i<t} = (f_{\bar{t}}^{q})_{\bar{t}=1, q=1}^{t-1, Q}.$$

A general Seq2Seq network is effective for multiple time series problems that rely on a known training output size which is often not the case.  However, if the natural sequence size is not known a priori, a basic Seq2Seq is insufficient.  We create a model that is able to learn from training data that contains a capped output length during the training phase and predicts a dynamic output sequence during the prediction phase.  

\subsection{Dynamic Prediction Length Sequence to Sequence}

To create a model that predicts a dynamic output length for each time series, we introduce a function, $g(f_{i\le t})$ that captures confidence of the prediction at time $t$.  Note that the confidence function may be dependent on previous values $f_{1}, f_{2}, ...f_t$.  To determine output length, we measure the confidence function against a threshold value $\tau$ (which is a hyper-parameter) resulting in 

\begin{equation}
\min_\theta E_{(X,Y)} \sum_{q=1}^Q \sum_{g(f_{i\le t}^{q}(X;\theta;f_{i<t}))\ge \tau} KL(Y_{t}^{q} || f_{t}^{q} (X;\theta; f_{i<t} )).
\label{originaleqn}
\end{equation}

For the remainder of our work, we let $L_{t}^{q} := KL(Y_{t}^{q} || f_{t}^{q} (X;\theta; f_{i<t}))$ and 

\noindent $G_{t}^{q} := g(f_{i\le t}^{q}(X;\theta;f_{i<t}))$.  Note that we use the output of the softmax function $f_{t}^{q} (X;\theta;f_{i})$ as a loss measure in the Kullback-Leibler divergence and the confidence measure for dynamic output length.  As an example, $G_{t}^{q}$ can return the maximum predicted probability at time $t$, but it can also be total variation between one time step and the next.

The baseline loss function above contains a flaw.  Since the set of optimal values are $\Theta^* := \{\theta^* : G_{t}^{q} (\theta^* ) \ge \tau\}$, the model does not have to make accurate predictions to minimize the loss function.  It can for example select $\theta$ such that $G_{1}^{q} <\tau$ and thus an ``empty" sum.  Therefore, we add a penalty to the loss function that punishes the model for not meeting the threshold to generate accurate prediction outputs.  

Next, we introduce a second hyperparameter, $\lambda$, with the new penalty function in the loss.  If the model makes few predictions, we can increase $\lambda$ and decrease $\tau$.  To balance the F1 score and output length, we adjust each of these hyperparameters.  By using the indicator function $I[G_{t}^{q} \ge \tau]$ for $1$ if $G_{t}^{q} \ge \tau$ and $0$ otherwise, the loss function reads

\begin{equation}
\min_\theta E_{(X,Y))} \sum_{q=1}^{Q} \sum_{t=1}^T \big{(} I[G_{t}^{q} \ge \tau] L_{t}^{q} + \lambda \max(\tau - G_{t}^{q} ,0) \big{)}.
\label{indicator}
\end{equation}

\noindent By design, if the model masks the Kullback-Leibler divergence (i.e. $G_{t}^{q} < \tau$), the penalty function produces a nonnegative output and vice-versa. 

In addition to designing our penalty function to counter the Kullback-Leibler error, we make the penalty function in the form of a rectifier unit. We do this since rectifier units are effective with neural networks for training and have a simple non-vanishing gradient.  Since there is only a single gradient difference between this function and the Kullback-Leibler loss, we save a lot of computational effort.

We also consider a smooth masking function to improve differentiability.  We propose a form of the sigmoid function.  With the sigmoid function our loss reads   

\begin{equation}
\min_\theta E_{(X,Y)} \sum_{q=1}^{Q} \sum_{t=1}^{T} \big{[}H_{t}^{k,\tau}(G_{t}^{q} ) L_{t}^{q} + \lambda \max(\tau-G_{t}^{q},0)\big{]}.
\label{sigmoid}
\end{equation}

\noindent where

\begin{center}
$H_{t}^{k,\tau} (x) = \frac{1}{1+e^{-k(\frac{x-\tau}{1-\tau})}}$
\end{center}

\noindent for positive and possibly large hyperparameter $k$.  Since the loss function  in (\ref{sigmoid}) is differentiable, standard back propagation can be used.  On the other hand, (\ref{indicator}) is not differentiable since the number of summation term depends on trainable parameters $\theta$. 

Up until this point, we have not considered the various possibilities for the function

\noindent $g(f_{i\le t}^{q}(X;\theta;f_{i<t})$.  To properly balance accuracy and future predictions, we require that the function can express the confidence of the model and/or be an effective measure of volatility.  We choose four different functions that range in both complexity and time dependency:

\begin{itemize}
\item Maximum: $\max_j f_{t,j}^{q}(X;\theta;f_{i<t})$
\item Confidence Distance: $\max_j (f_{t,j}^{q}(X;\theta;f_{i<t}))-\max_{k, k\ne j^*} (f_{t,k}^{q}(X;\theta;f_{i<t}))$ where $j^* := \argmax_j (f_{t,j}^{q}(X;\theta;f_{i<t}))$
\item Total Variation: $\max_j |f_{t,j}^{q}(X;\theta;f_{i<t}) - f_{t-1,j}^{q}(X;\theta;f_{i<t-1})|$
\item Wasserstein/Earth Mover Distance: $W^1 (f_{t}^{q}(X;\theta;f_{i<t}),f_{t-1}^{q}(X;\theta;f_{i<t-1}))$.
\end{itemize}

The first two functions are based upon the current prediction time while the other two are dependent on the current and previous outputs.  Confidence distance and maximum measure confidence of the current prediction.  If the model is struggling between two different labels (uncertainty), then we expect this value to be small enough to be below the threshold.  Both Total Variation and the Wasserstein Distance are classic measures for the distance between two probability distributions.  Note that we specifically choose the first Wasserstein Distance, which is also known as the Earth Mover Distance.  In addition, because total variation and the earth mover distance measure volatility, we define $G_{t}^{q}$ to be the negative of these two measures, which then leads to $G_{t}^{q} \le \tau$.

\subsection{Training and Inference}

There are differences between the training of a typical Seq2Seq model and our new model during training and prediction.  For example, when our network stops at $\hat{T}(q)$ for time series $q$, we no longer consider the accuracy error terms for $t>\hat{T}(q)$ when computing the gradient.  

In the forward pass, we stop when $G_{\bar{t}}^{q}< \tau $ occurs the first time for time series $q$ and thus $\bar{t}=\bar{t}(q)$.  It may happen that for some $G_{\hat{t}(q)}^{q} \ge \tau$ for some $\hat{t}(q) > \bar{t}(q)$.  When this occurs, we do not consider the terms corresponding to $\hat{t}(q)$ in the Kullback-Leibler divergence or in our F1 score calculation.  With the indicator function, the outputs from the Kullback-Leibler divergence are completely masked to zero.  However, with the sigmoid function, the outputs below $\tau$ are only partially masked.  In addition, when utilizing the sigmoid function, we calculate the F1 scores in the same manner as in the case of the indicator function.

After obtaining $\bar{t}(q)$ in the forward pass for each time series $q$, we change the loss to the following when using (\ref{originaleqn}):

$$\min_\theta E_{(X,Y))} \sum_{q=1}^{Q}\left[  \sum_{t=1}^{\bar{t}(q)} L_{t}^{q} + \sum_{\hat{t}=\bar{t}(q)+1}^{T} \lambda \max(\tau - G_{\hat{t}}^{q} ,0) \right].$$

From this point on, we do standard backpropagation treating $\bar{t}(q)$ as fixed for the current sample.

\section{Computational Study}

\subsection{Data Preparation}
We study our model on financial time series data.  Our data consists of fourteen years of securities at five minute tick intervals for two sets consisting of twenty-two ETF's and five distinct commodities.  The identity of the securities are unknown; therefore, we do not incorporate any additional features such as news and market announcements.  To clarify, the dataset consists of only the single tick price rather than prices on multiple levels of an order book.  In addition, we do not have trade volume information.  The prices for the twenty-two ETF dataset are the returns of each security, which is the relative change in price from one 5 minute interval to the next.  The five commodity dataset consists of the price of each security at each time $t$.  In order to have consistency between the two datasets, we choose to only consider the returns of the five commodity dataset. 

We create two representations of our data: one for the feed forward and seq2seq networks and the other for the convolutional seq2seq network.  First, we create a sequence of inputs consisting of $\bar{T}$ returns for each financial asset.  Since our data can be viewed as a continuous streaming sequence (no obvious beginning or end), we create sequences of various size $\bar{T}$.  For each sequence, the next sequence moves forward by a 5 minute interval.  We found that including this overlap increased prediction accuracy.  In our best performing feed forward networks, our feature vectors consist of ten returns.  For a seq2seq network we increase the sequence size to twenty.

Since an increase in the sequence size greatly increases the computation time of a seq2seq neural network, we consider a 1-dimensional convolutional LSTM.  A convolutional LSTM layer can take much larger sequences because the convolutional step in each LSTM node reduces the total number of sequences via a convolutional kernel.  Since we input multiple returns from different financial assets, each channel of convolution represents a difference financial security (similar to RGB).  We tried two models: convolution feeding into LSTM and ConvLSTM (\citet{xingjian2015convolutional}).  All seq2seq models are embedded with our loss function unless stated otherwise.

For normalization, we use the standard approach of calculating the mean and standard deviation of the training set and using that to normalize both the training and validation/test sets.  For our data, we use one year of training data (47,000 samples), and classify the next week of returns (1,000 samples).

We utilize a walk-forward methodology to evaluate our model over the entire dataset.  After each training phase we move the training and testing windows one week forward.  The first week of the previous training data is dropped form the training data of the next phase.  Each training is warm-started (pretrained) from the previous window.

To classify the returns of each security, we split the labels into five classes: large upward movement, small upward movement, insignificant movement, small downward movement, and large downward movement.

We calculate the mean ($\mu_D$) and standard deviation ($\sigma_D$) of the returns $ s_D $ of the previous day to create our labeling scheme of five classes for the returns of day $D+1$.  To classify the return $x_{t}^{D+1}$on day $D+1$ at time $t$, we use the following rules.  $$x_t^{D+1} < \mu_D - \sigma_D,$$
$$\mu_D - \sigma_D \le x_t^{D+1} < \mu_D - \beta \sigma_D,$$
$$\mu_D - \beta \sigma_D \le x_t^{D+1} < \mu_D +\beta \sigma_D, $$
$$\mu_D + \beta \sigma_D \le x_t^{D+1} < \mu_D + \sigma_D,$$
$$\mu_D + \sigma_D \le x_t^{D+1} $$

Here $\beta$ is a paramter.  We set $\beta$ such that roughly 50\% of all values lie within insignificant movement ($\beta=0.14$) for the twenty-two security dataset.  For the other dataset, a few of the securities contain large data imbalances at the large upward and downward movements.  Since this is the case, we pick $\beta=0.1$ so that the majority class contains roughly 50\% of all labels.  These choices lead to class imbalances of roughly [$5\%$, $20\%$, $50\%$, $20\%$, $5\%$].

\subsection{Results}

The following results are based upon the best sequence sizes, neurons per layer, optimization method, and number of layers found through hyperparameterization of each network.  The models were trained on Titan X's and Nvidia 1080's and implemented in tensorflow.  For each dataset, we have one model to classify all twenty-two ETF's from one dataset and one model to classify all five commodities in the other dataset.  We test the ETF dataset on all baseline and final model architectures.  Due to computational time constraints, we only calculate the results of the commodity dataset using the best model from the ETF dataset.

\subsection{Baseline Test}

For our first experiments, we use a FFN, LSTM network, LSTM Seq2Seq network, and a ConvLSTM Seq2Seq network in the setting of a fixed prediction of 1 or 10.  The FFN consists of two layers with sixty-four neurons and ten returns for each security.  We train the FFN with the stochastic gradient optimization method.  The basic recurrent LSTM network consists of two LSTM layers, each with sixty-four neurons and input sequence size of twenty.  For all recurrent networks, we use the ADAM optimization method by \citet{kingma2014adam}, which is known to perform well for recurrent networks.  We train for a number of epochs until the F1 score no longer increases on the validation dataset.  Last, the sequence to sequence network consists of one encoder and one decoder layer, each with sixty-four neurons and either LSTM or ConvLSTM layers.  We tried using deeper networks, but those resulted in much lower F1 score.  We utilize the same model architecture for both the ETF and commodity datasets.

We are using financial security datasets, which typically contain imbalanced classes.  When considering imbalanced classes, the traditional accuracy measure does not display the true effectiveness of a model.  Therefore, we utilize the F1 score, which accounts for imbalanced classes.  The F1 score is calculated via a one-against-all structure of precision and recall for each of the five classes.  The values we report are an average of the F1 scores for each class and security.  

The FFN, LSTM, LSTM Seq2Seq, and ConvLSTM Seq2Seq results are presented in Table \ref{basefin}.  To produce these results, we give each model a warm start by training on 5 windows of data.  We then train the models for 25 more windows on the 22 ETF's and 5 commodities datasets.  The F1 scores in Table \ref{basefin} are the average test scores over this 25 window period.

\begin{table}
\captionof{table}{F1 Scores of Baseline Models} \label{tab:title} 
\begin{tabular}{lrr}
\hline
Architecture    & ETF's & Commodities \\
\hline
FFN (One Pred)    &  0.176    & 0.159    \\
LSTM (One Pred)    & 0.209    & 0.286         \\
ConvLSTM (One Pred) & 0.513 & 0.410  \\
LSTM Seq2Seq (One Pred)  & 0.598     & 0.513   \\
LSTM Seq2Seq (Ten Pred)  & 0.509     & 0.474  \\
\lasthline
\end{tabular}\par
\label{basefin}
\end{table}

We use our baseline results in Table \ref{basefin} to determine the best architecture for our dynamic output length model and to examine the differences of the various seq2seq networks.  We observe that each single prediction model generally increases in performance when using more robust models (such as seq2seq).  As expected, the seq2seq models and recurrent LSTM models far outperform the FFN, but we did not expect such a large performance increase when utilizing a seq2seq model.  Even when making 10 predictions forward in time, the seq2seq model more than doubles the F1 score compared to the traditional LSTM network that makes a single prediction.  With ConvLSTM layers, we found that the best formulation is utilizing 1D convolutions with each security representing a channel in the image.  Ultimately, we find that the best ConvLSTM seq2seq model does not perform as well as the LSTM seq2seq model.  Since the LSTM seq2seq model is by far the most accurate, we choose it for our dynamic output length model.

\subsection{Maximum and Confidence Distance Confidence Functions}

We begin with the two confidence functions that depend only upon the current prediction: maximum and CD.  We study the sensitivity of $\tau$ and  $\lambda$, and the comparison between the indicator (Ind) and sigmoid (Sig) functions, i.e. loss functions (4.2) and (4.3) respectively.  In the following sections, we refer to the indicator and sigmoid functions as masking functions because they mask the output from the Kullback-Leibler divergence.

\begin{figure}[ht]
\centering
\begin{minipage}[b]{0.49\textwidth}
\includegraphics[width=\textwidth]{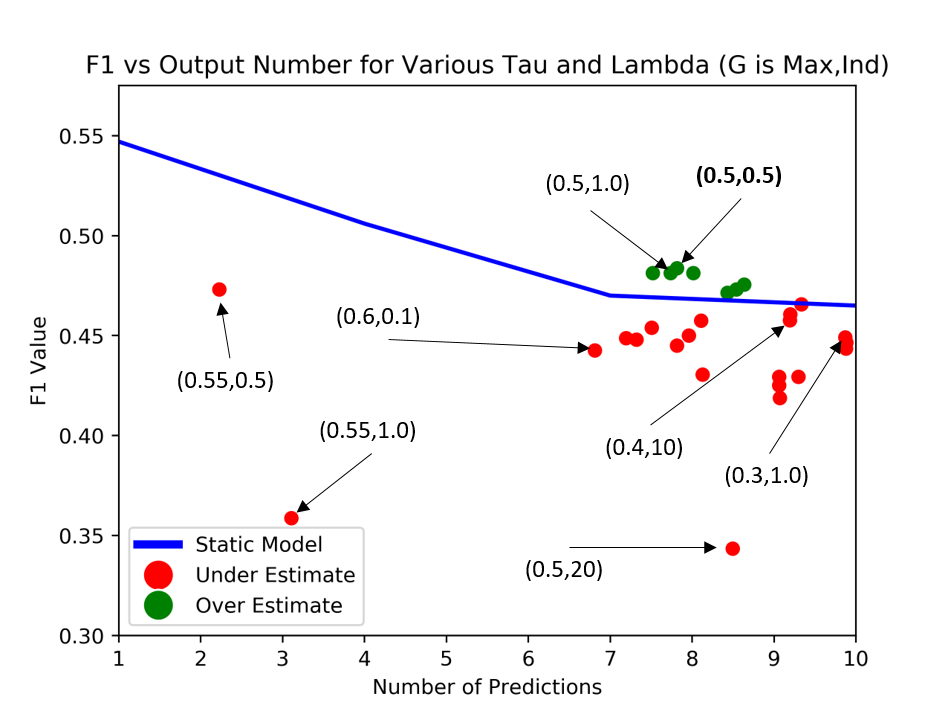}
\end{minipage}
\begin{minipage}[b]{0.49\textwidth}
\includegraphics[width=\textwidth]{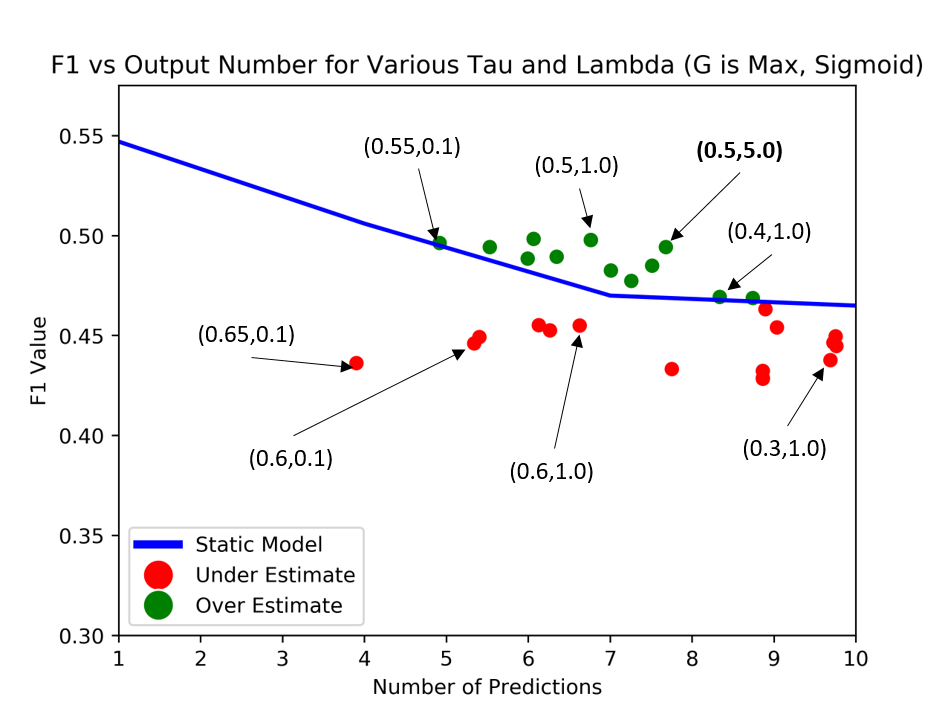}
\end{minipage}
\caption{F1 and average output length for pairs of $\tau$ and $\lambda$.  We use the maximum confidence function with the indicator function on the left and sigmoid on the right.}
\label{gmaxall}
\end{figure}

A large gradient for $\lambda\max(\tau-G_{t}^{q},0)$ means that we encourage the model to make more predictions.  We can control the output length by adjusting the hyperparameters $\lambda$ and $\tau$.  In Figure \ref{gmaxall} we present a range of $\tau$ and $\lambda$ values and the respective F1 scores with $G_{t}^{q}$ being the maximum.  In addition, we place an annotation of the best pair $(\tau, \lambda)$ that results in the largest distance above the static curve in bold.  We add this annotation to all figures of this type.

In Figure \ref{gmaxall}, we first create a line that gives the F1 scores from a static seq2seq model for 5 training windows after a 5 window warm start.  We compute F1 for $1, 4, 7, 10$ predictions by the static model and interpolate the remaining values.  It is important to point out that the F1 scores here are different than in Table \ref{basefin} since we are measuring 5 windows rather than 25.  We expect the dynamic model to be above this curve when using appropriate values for $(\tau, \lambda)$.  Each point in Figure \ref{gmaxall} is the average output length and F1 score of the 22 ETF dataset for a pair $(\tau,\lambda)$.  On the left is the indicator masking function and the right is the sigmoid masking function.  For both figures, we use the same values for $\tau$ and $\lambda$.  We find that with $\tau = 0.5$, our model has high accuracy with a relatively large average output length.

The red points are a pair $(\tau,\lambda)$ that perform worse than the static model and the green are those that perform better.  We expect to find many pairs of hyperparameters that lead to superior F1 scores since our model uses a dynamic prediction length for each sample.  There are a few outliers that fall well below the estimated performance in the figure on the left, which uses the indicator masking function.  The sigmoid function on the right has many more points above the static curve and does not have any points below the $0.4$ F1 score.  We observe that many red points usually have large $\lambda$'s.  When $\lambda$ is too big, the model stops training for prediction accuracy and only focuses on creating more predictions.  Note that not all large $\lambda$'s lead to strictly poor performance since the best sigmoid pair has $\lambda = 5.0$.
 
\begin{figure}[ht]
\centering
\begin{minipage}[b]{0.49\textwidth}
\includegraphics[width=\textwidth]{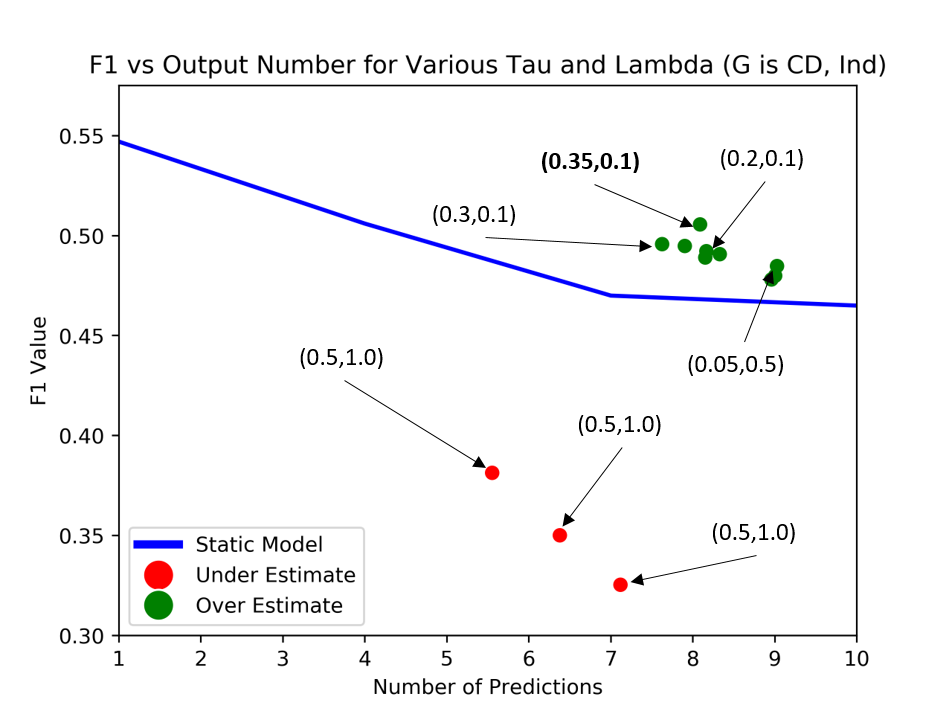}
\end{minipage}
\begin{minipage}[b]{0.49\textwidth}
\includegraphics[width=\textwidth]{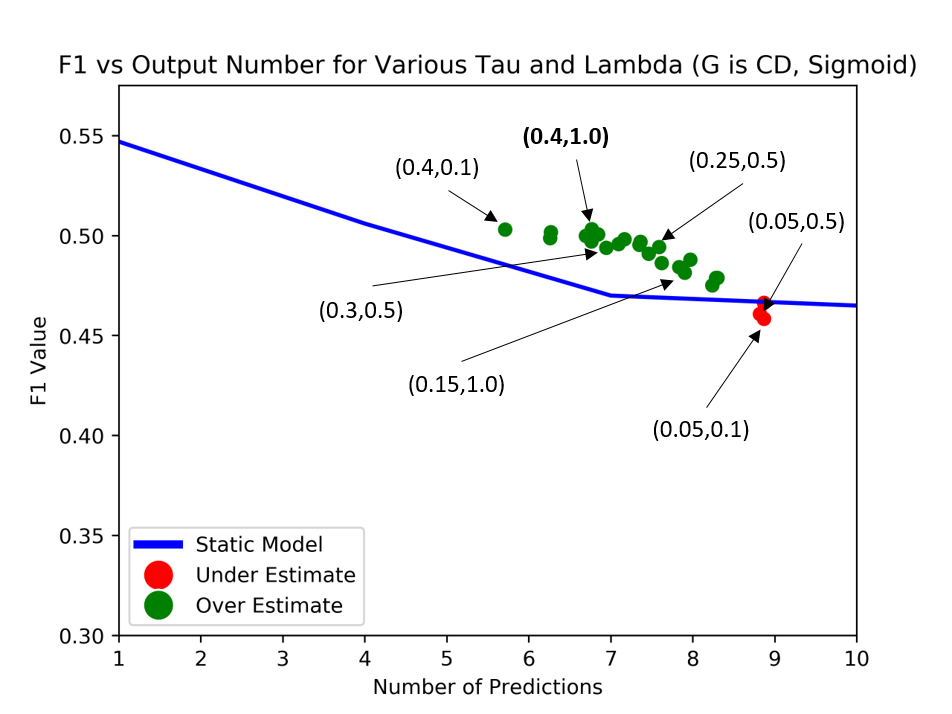}
\end{minipage}
\caption{F1 score and average output length for pairs of $\tau$ and $\lambda$.  We use CD with the indicator function on the left and sigmoid function on the right.}
\label{cdall}
\end{figure}

Next we consider  CD.   As in Figure \ref{gmaxall}, Figure \ref{cdall} shows a hyperparameter search $(\tau,\lambda)$ with the indicator masking function on the left and sigmoid masking function on the right.  Note that with CD, the number of points below the static line is smaller compared to the maximum function.  When $\lambda$ is too large, the indicator version has three points that fall far below our estimation.  It is important to point out that there is a large distance between the green points and the blue curve that is rarely seen with other confidence functions.  Based on these results, we recommend using the CD over the maximum function.  It requires very little extra computation and leads to better F1 with a similar average output length.

\begin{figure}[ht]
\centering
\begin{minipage}[b]{0.49\textwidth}
\includegraphics[width=\textwidth]{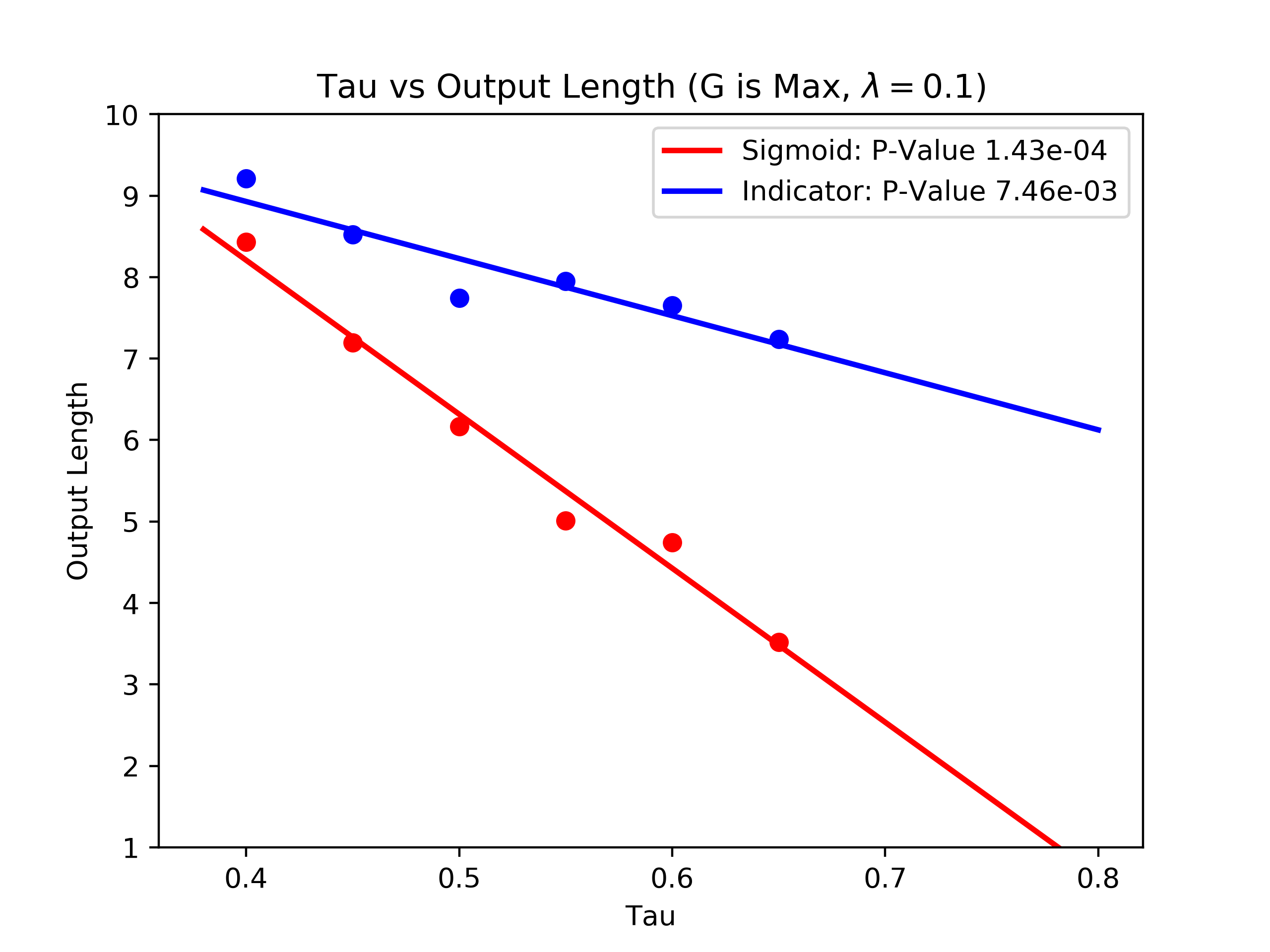}
\end{minipage}
\begin{minipage}[b]{0.49\textwidth}
\includegraphics[width=\textwidth]{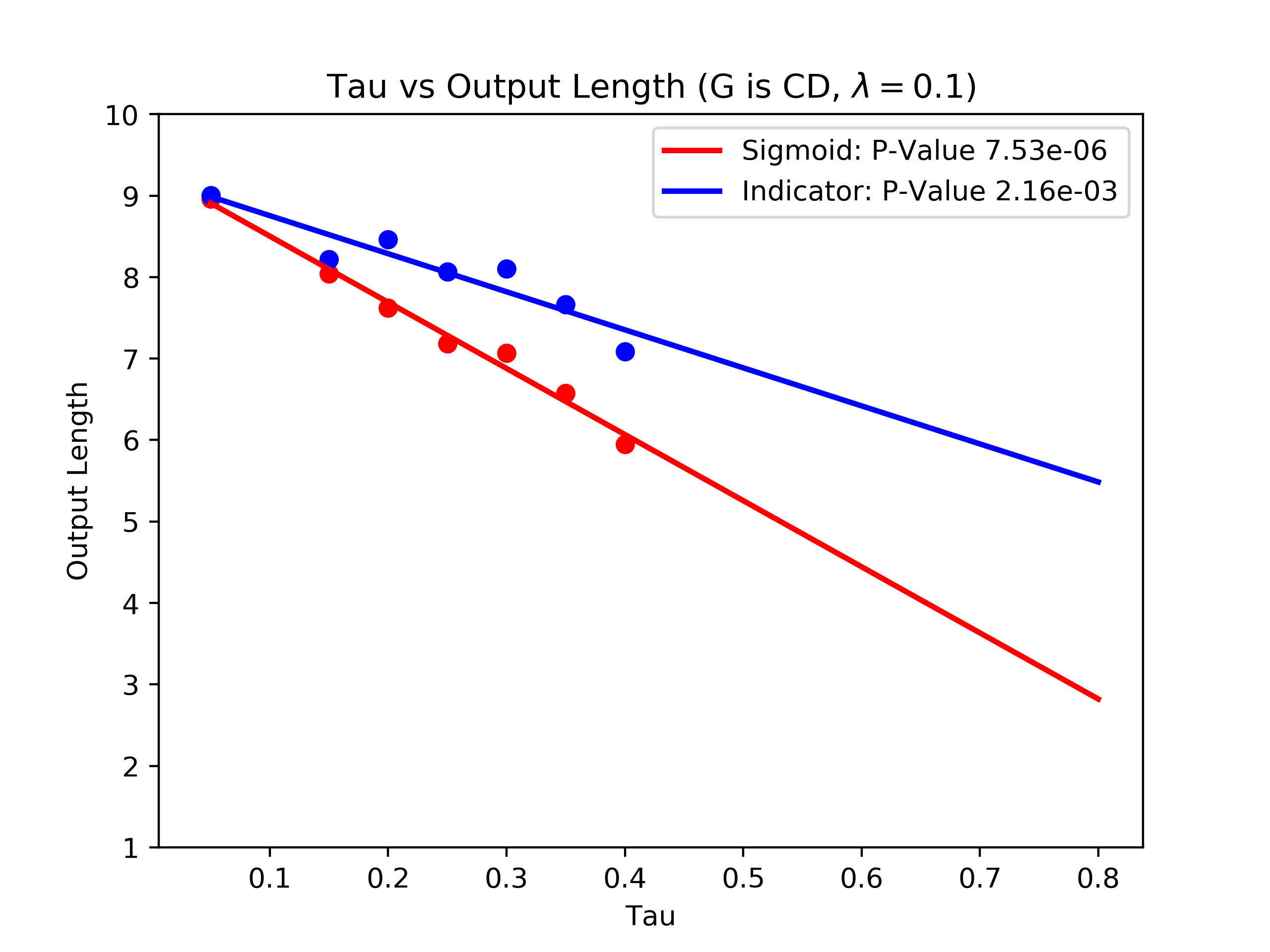}
\end{minipage}
\caption{The output length and $\tau$ values for both CD and maximum functions with $\lambda=0.1$.  The trend lines are made via linear regression and show high statistical significance between $\tau$ and output length.}
\label{sensemaxcd}
\end{figure}

In Figures \ref{gmaxall} and \ref{cdall}, we show that large $\lambda$'s usually lead to inferior models.  Therefore, we explore $\tau$ to see if we can find a relationship between this hyperparameter and the average output length.  We present two images in Figure \ref{sensemaxcd} of the sensitivity of the indicator and sigmoid masking functions with maximum and CD functions.  We create these figures by choosing $\lambda=0.1$, which produces the best performance for all cases of $\tau$ and plot the relationship between average number of predictions and $\tau$.  Note that we use the same approach of utilizing a 5 window warm start and measure the average output length based upon the next 5 windows. 

On the left, we observe that the slope for the sigmoid masking function is sharper than the indcator function for both confidence functions.  In the case of these two functions, the sensitivity is not nearly as large as the other two confidence functions that we explore in the next section.  For CD, the slope is much smaller (in terms of absolute value), which provides additional evidence to recommend it over the maximum confidence function.

Given these two confidence and masking functions, we recommend using CD with the sigmoid masking function.  Although the sigmoid masking function is more sensitive to $\tau$, we find that sigmoid does not have as many points falling under the static curve.  Next, we examine the two confidence functions that depend on the current and previous predictions.

\subsection{Total Variation and Wasserstein Confidence Functions}

\begin{figure}[ht]
\centering
\begin{minipage}[b]{0.49\textwidth}
\includegraphics[width=\textwidth]{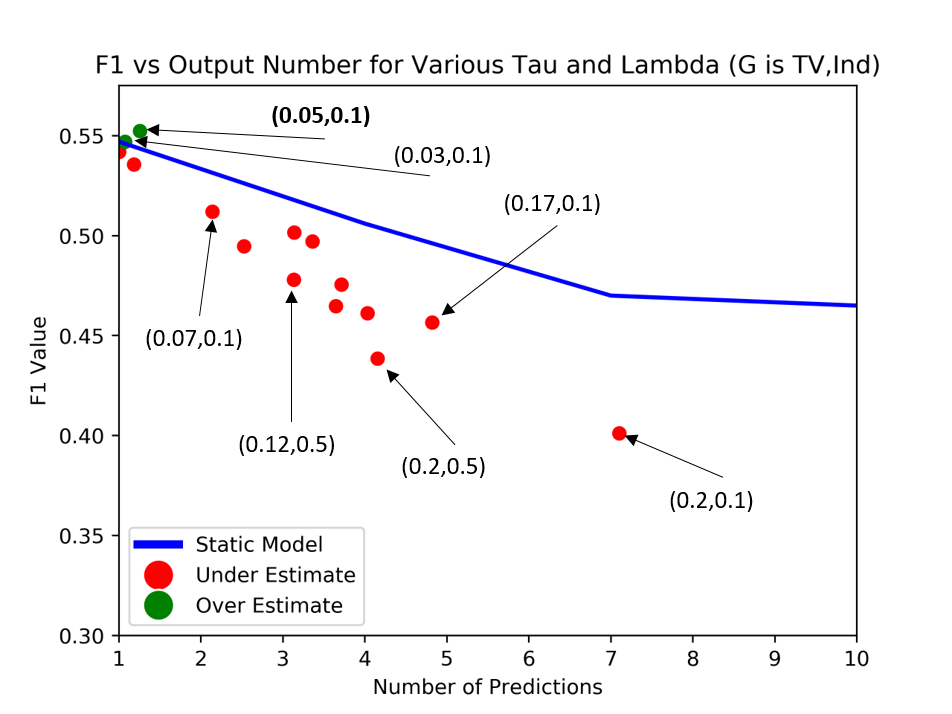}
\end{minipage}
\begin{minipage}[b]{0.49\textwidth}
\includegraphics[width=\textwidth]{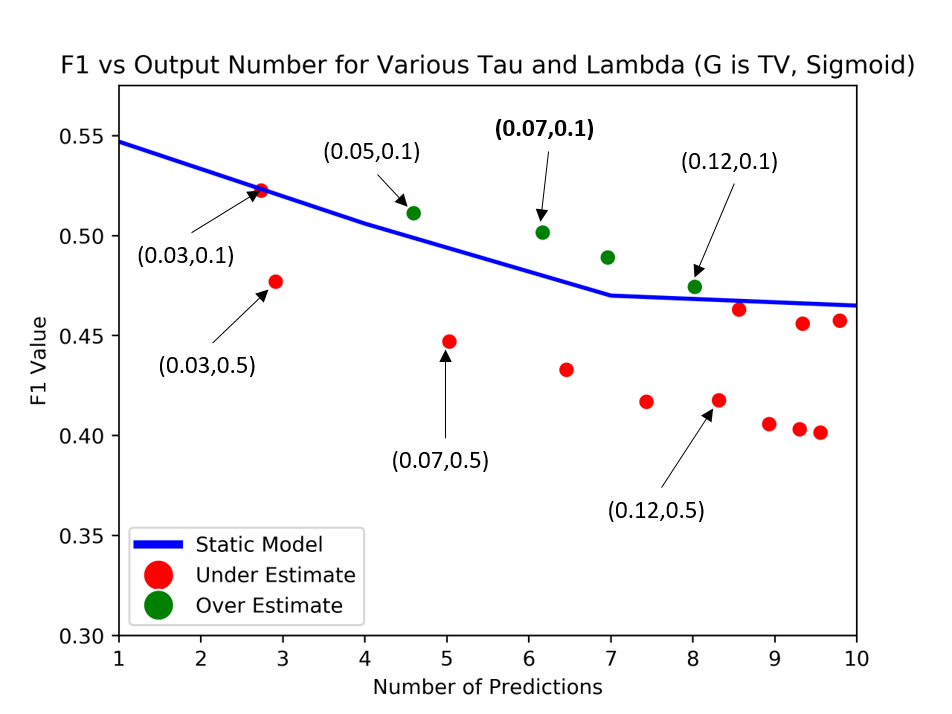}
\end{minipage}
\caption{The F1 score and respective output length for various $(\tau,\lambda)$ pairs with TV.  The indicator function is on the left while the sigmoid function is on the right.}
\label{tvall}
\end{figure}

The other two functions we test are total variation (TV) and the first Wasserstein distance (EMD).  We design the first two functions, maximum and CD, to find the confidence of the current prediction using only the current prediction and input sequence.  On the other hand, we use TV and EMD to determine the volatility between the current and previous prediction.  If the volatility is low, we expect the model to be confident enough to make several predictions.  We first examine the results from TV in Figure \ref{tvall}.

TV does not garner high performance with either masking function.  There are some points with the sigmoid masking function that are slightly better than the expected F1 scores, but TV is clearly not a successful measure.  There are likely some cases where the TV may be large between two predictions, but the model may still be confident about that prediction.  For example, if the prediction changes from one label to another at the next prediction step, the TV between these two predictions may be large. However, this does not necessarily imply that the model is not confident about its next prediction.  Also note that there are two rows of points above and below the static curve of the sigmoid version.  The difference between these two is that the value for $\lambda$ is larger for the points below the estimated curve, which is similar to the results we obtain form the maximum and CD functions.  

\begin{figure}[ht]
\centering
\begin{minipage}[b]{0.49\textwidth}
\includegraphics[width=\textwidth]{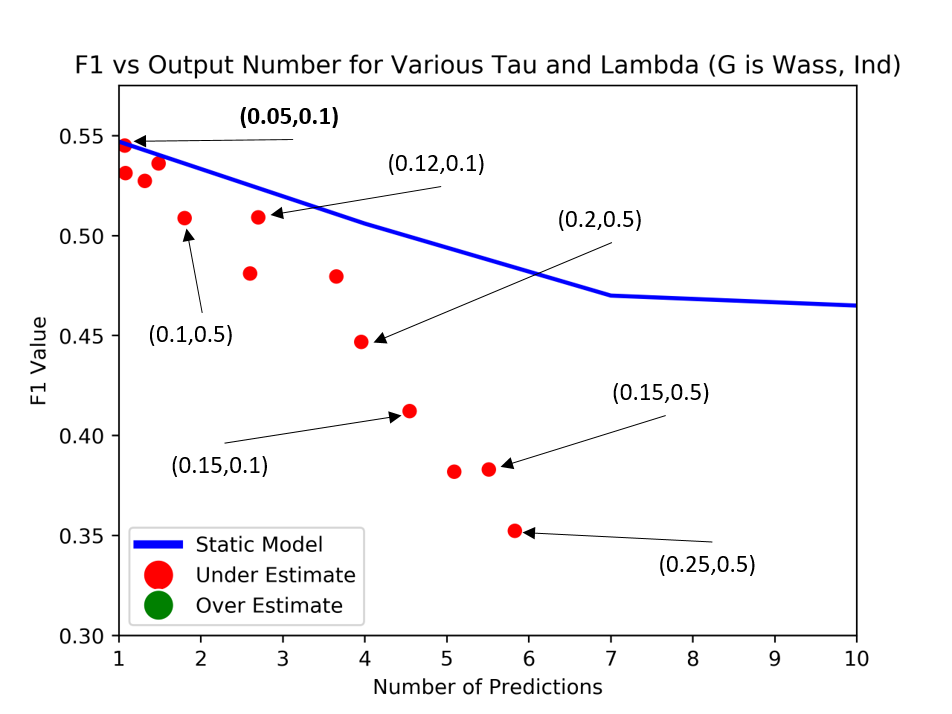}
\end{minipage}
\begin{minipage}[b]{0.49\textwidth}
\includegraphics[width=\textwidth]{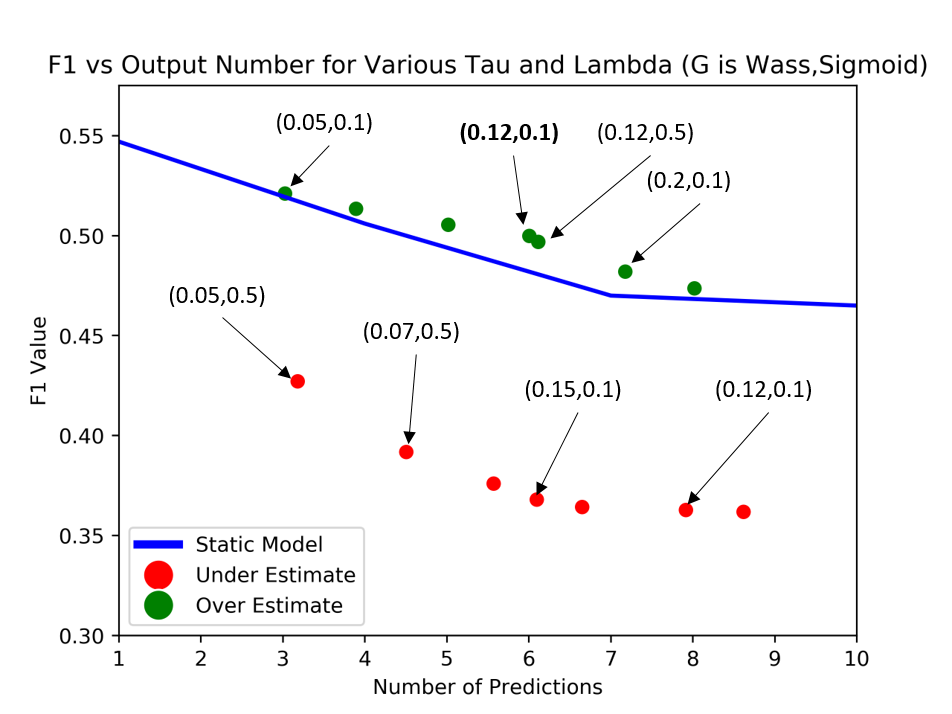}
\end{minipage}
\caption{The F1 score and respective output length for various $(\tau,\lambda)$ pairs with EMD.  The indicator function is on the left while the sigmoid function is on the right.}
\label{wassall}
\end{figure}

In Figure \ref{wassall}, we observe that EMD contains a similar pattern to TV.  Similar to TV, the indicator masking function with EMD performs poorly on the left compared to the better F1 scores of the sigmoid masking function on the right.  EMD performs slightly better than TV with a few more points above the estimated curve.  This likely occurs because EMD is a more robust probability measure.  Instead of finding the maximum distance between two probability measures $P$ and $Q$ (TV), EMD finds the cost of transforming the entirety of one distribution $P$ into another distribution $Q$.  EMD is more robust, but a label change from one time step to the next may correlate with a large EMD.  However, this does not always imply lower prediction confidence. 
 
\begin{figure}[ht]
\centering
\begin{minipage}[b]{0.49\textwidth}
\includegraphics[width=\textwidth]{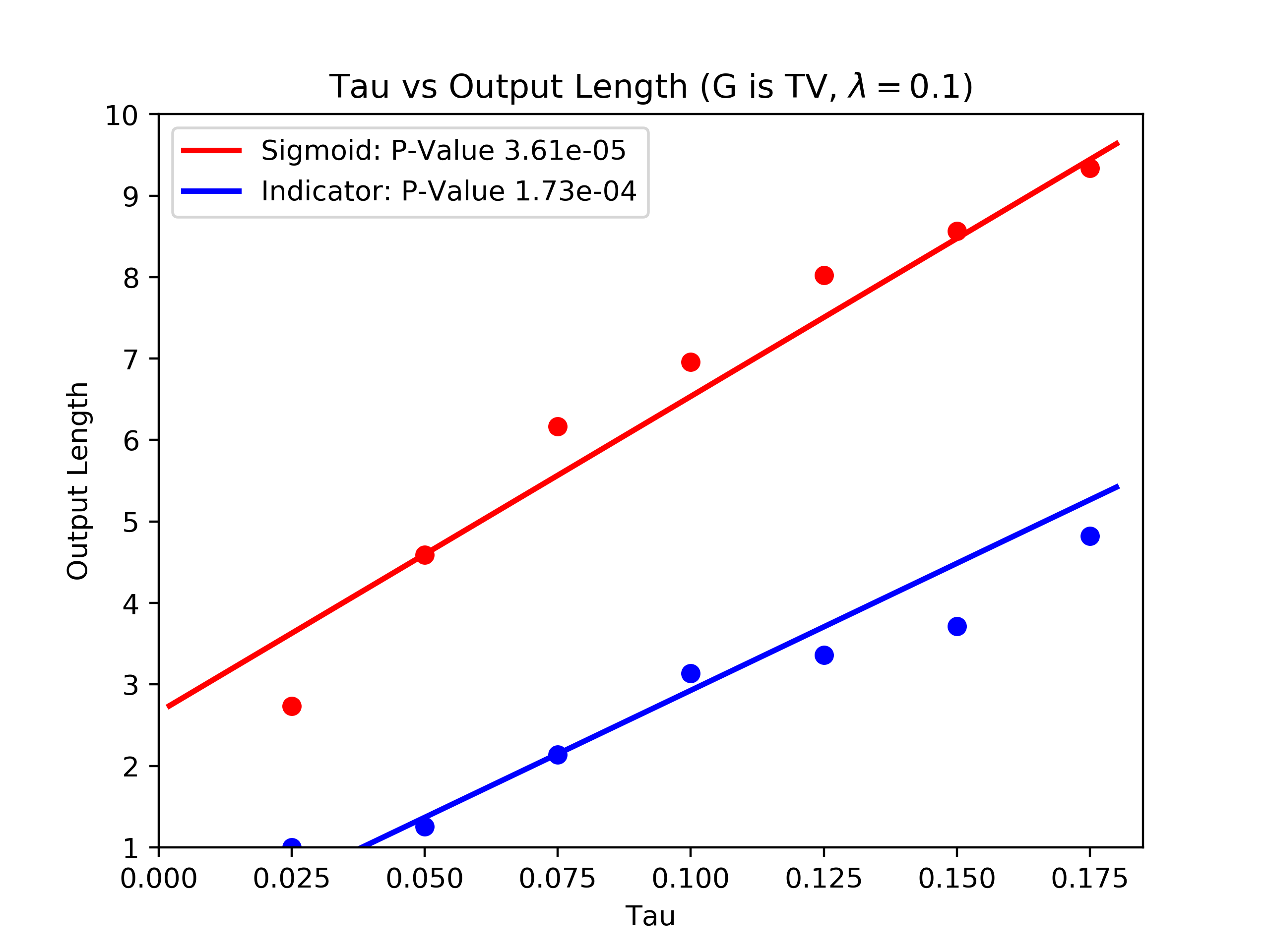}
\end{minipage}
\begin{minipage}[b]{0.49\textwidth}
\includegraphics[width=\textwidth]{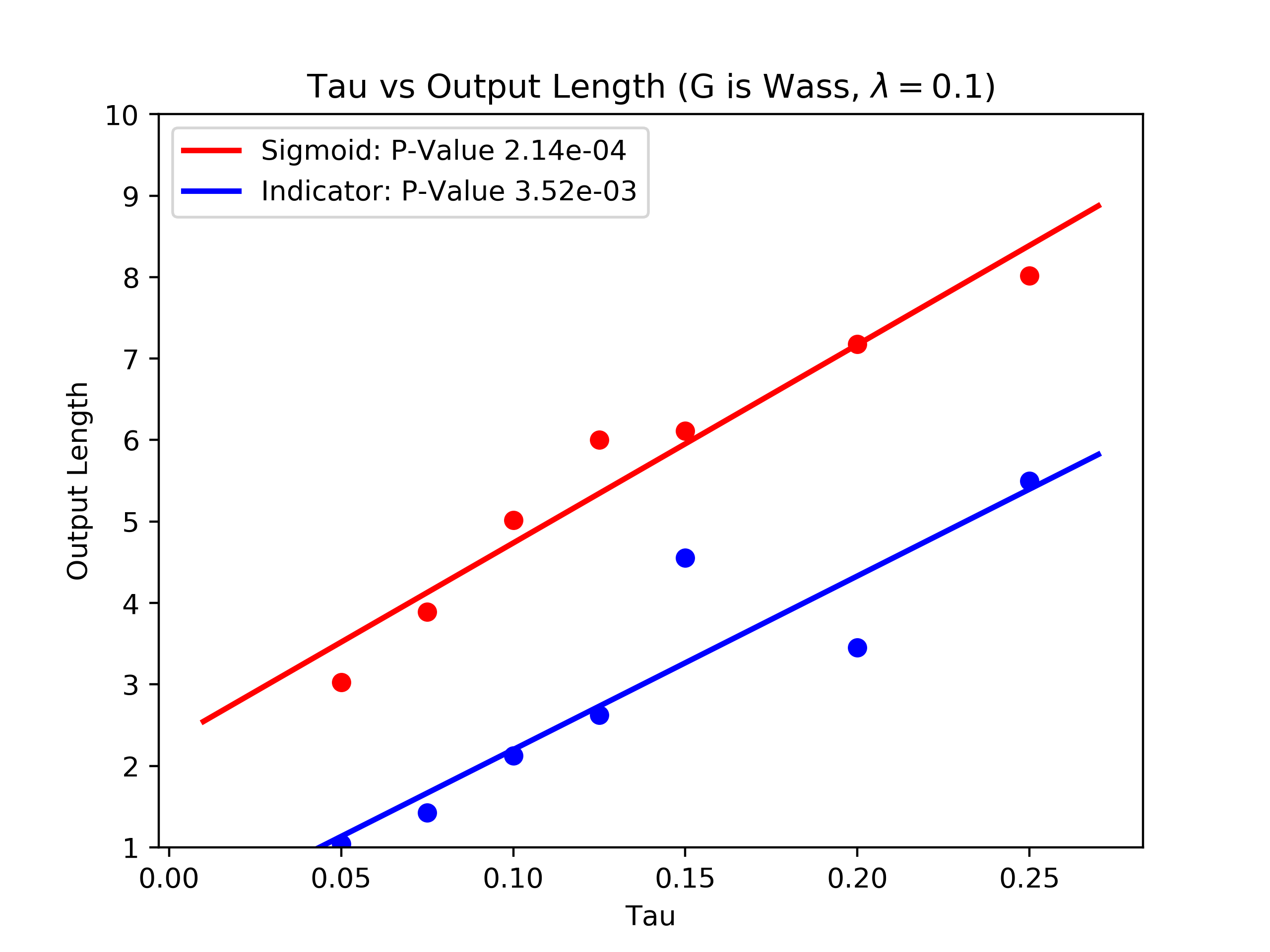}
\end{minipage}
\caption{The output length and $\tau$ values for both TV and the EMD with $\lambda=0.1$.  The trend lines are made via traditional linear regression and show high statistical significance between $\tau$ and output length.}
\label{sensetvwass}
\end{figure}

To conclude our examination of TV and EMD, we observe the relationship between the number of predictions and $\tau$ in Figure \ref{sensetvwass}.  Note that the relationship is the opposite of the maximum and CD functions since we reverse the relationship with respect to $\tau$, which we clarify in Section 4.3.1. 

The slopes for the sigmoid versions of both metrics are larger than their indicator function counterparts.  We also observe that the slopes are large overall when compared to both the maximum and CD confidence functions.  Slight changes in $\tau$ lead to large changes in output lengths for TV and EMD.  In general, we found that the values of these two metrics tend to be small, which is what leads to their sensitivity to $\tau$.  Overall, for all metrics, the sigmoid masking function slope is larger in magnitude than the same version with the indicator masking function.  Also, TV and EMD show that confidence in predictions depends upon more than volatility.

\subsection{Five Commodity Results}

\begin{figure}[!ht]
\centering
\includegraphics[width=0.7\textwidth]{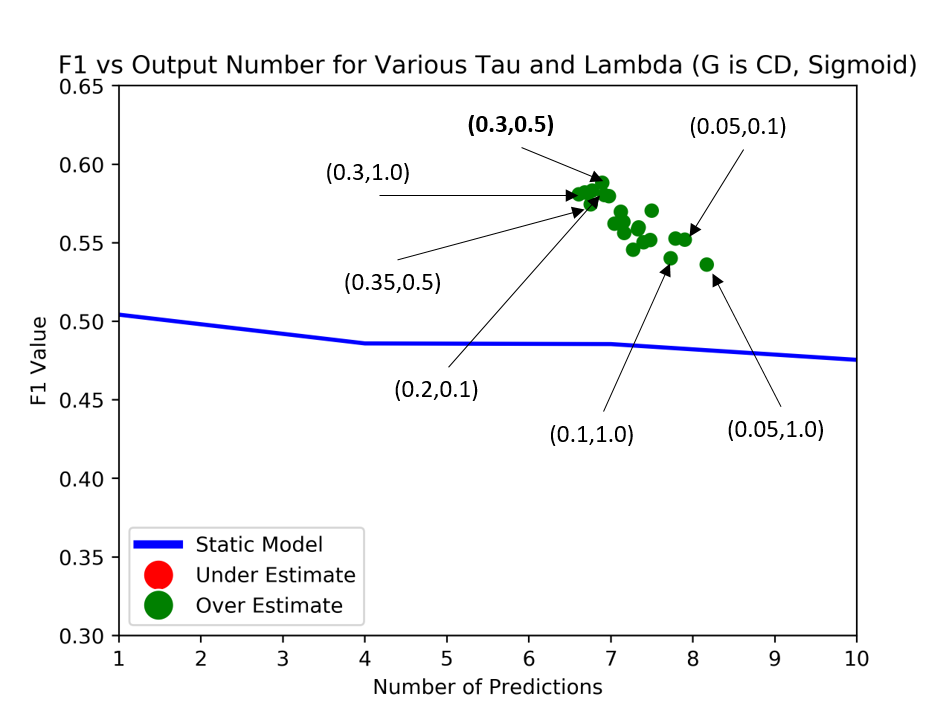}
\caption{The F1 score and respective output length for various $(\tau,\lambda)$ pairs with CD and the sigmoid masking function with the five commodity dataset.}
\label{sigcdcom}
\end{figure}

Next, we test our model on the five commodity dataset.  Since we observe that the CD with the sigmoid masking function provide the best performing model with the ETF dataset, we use the identical functions on the commodity dataset.  In addition, we create four static models predicting $1, 4, 7, 10$ time steps each to create the blue curve as with the previous ETF figures.  In Figure \ref{sigcdcom}, we present the dynamic model results with the commodity dataset.  As with the previous figures of the same type, we run the model for a total of 10 walk-forward steps and average the final 5 F1 scores, which are reported in Figure \ref{sigcdcom}.

\begin{figure}[!ht]
\centering
\includegraphics[width=0.7\textwidth]{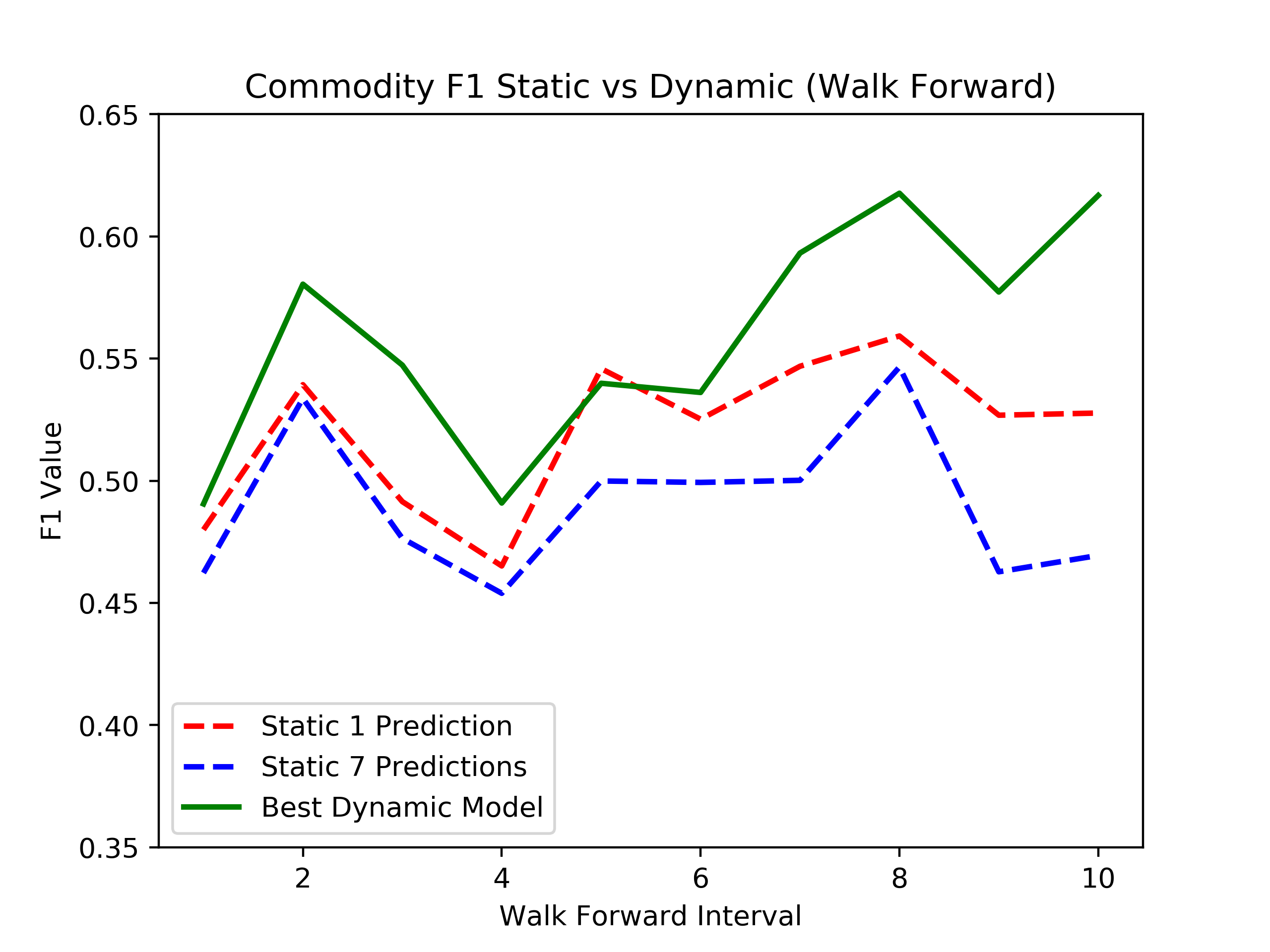}
\caption{The F1 score of our best dynamice model: $\tau = 0.3$ and $\lambda=0.1$ and an average prediction length of 7.  We showcase its F1 score for each walk forward period against two static models.}
\label{sigint}
\end{figure}

The first thing we observe is that this is the first time that every single pair $(\tau, \lambda)$ is above the static predictions.  This probably occurs because of volatility in the commodity dataset.  Some commodities are extremely volatile, and the price ranges from big highs to lows for a high percentage of the labels.  In addition, other commodities in the dataset have very low volatility for a long period of time.  Because of the skewed nature of the commodity dataset, it is possible for our model to only choose to predict at times when it can make many correct predictions.  To observe the dynamic model in Figure \ref{sigint}, we observe the F1 scores of our best dynamic commodity model and two static models over 10 walk forward periods.

The F1 score of our dynamic model is superior at nearly every walk forward time period in comparison to a static single prediction model, except for walk forward period 5.  As expected, the single prediction static model has a better F1 score on average than the static model making 7 predictions.  However, the difference between 1 prediction and 7 predictions is small, which is in stark contrast to the ETF dataset.  This provides additional evidence that the price volatility of the labels makes a static number of predictions difficult, and shows that a dynamic model can be advantageous with a time series dataset that has a skewed distribution.

\subsection{Summary of Results}

In Table \ref{sumdyn}, we present a summary of the best results for our dynamic model.  To measure which pair $(\tau, \lambda)$ is best, we measure the relative improvement between the F1 score of the dynamic model and the equivalent prediction length F1 score of the static model (from the blue curves shown in previous figures).  The dynamic model with CD and sigmoid does extremely well for the commodity dataset.  The same functions also give the best F1 score with the ETF dataset.

\begin{minipage}{\linewidth}
\centering
\captionof{table}{Dynamic Model Summary (All Sigmoid)} \label{tab:title} 
\begin{tabular}{lrrr}
\hline
Architecture    & F1 Gap \% & $(\tau, \lambda)$ & Prediction Length  \\
\hline
Dynamic Max (ETF)    &    5.44    & $(0.50, 5.0)$  & 7.68  \\
Dynamic CD (ETF)    &  6.45    & $(0.40, 1.0)$  & 6.76      \\
Dynamic TV (ETF) & 4.49 & $(0.07, 0.1)$ & 6.16 \\
Dynamic EMD (ETF)  &  3.72     & $(0.12, 0.1)$ & 6.00  \\
Dynamic CD (Commodity)  &  21.2     & $(0.30, 0.5)$  & 6.89\\
\lasthline
\end{tabular}\par
\label{sumdyn}
\end{minipage}

\section{Summary}

In this work, we create a new loss function with the seq2seq network that makes a dynamic number of predictions for each input sequence.  In addition, we construct a new metric called ``Confidence Distance" to measure the confidence the model has when making a prediction.  When testing the new model, we find that a dynamic prediction length model can outperform a similar static seq2seq network.  In addition, we find that of our four confidence functions, CD gives the most accurate predictions over maximum, TV, and the EMD.  We examine two versions of our model with each confidence function: masking the Kullback-Leibler divergence with the indicator function and masking with the sigmoid function.  We find that the sigmoid function leads to better and more reliable prediction F1 even though it is more sensitive to hyperparameter $\tau$.

When using the dynamic model, we recommend to keep the value for $\lambda$ low and to vary $\tau$ to get the desired output length to F1 ratio.  Seq2Seq models perform well with financial security data, and we recommend their use and to make the first decoder input the label associated with the final encoder input.  For the best dynamic output length performance, the best setting utilizes $\tau \in [0.05,0.5]$ and $\lambda \le 1.0$ with the confidence distance metric and sigmoid masking. 

\bibliographystyle{DeGruyter}
\bibliography{FinancePaper}

\end{document}